\documentclass{wscpaperproc}
\usepackage{latexsym}
\usepackage{graphicx}
\usepackage{mathptmx}
\usepackage{dsfont}
\usepackage{bm}
\usepackage{multirow}
\usepackage{microtype}

\usepackage{amsmath}
\usepackage{amsfonts}
\usepackage{amssymb}
\usepackage{amsbsy}
\usepackage{amsthm}
\usepackage{booktabs}

\usepackage[pdftex,colorlinks=true,urlcolor=blue,citecolor=black,anchorcolor=black,linkcolor=black]{hyperref}

\newtheoremstyle{wsc}
{3pt}
{3pt}
{}
{}
{\bf}
{}
{.5em}
{}

\theoremstyle{wsc}

\newtheorem{definition}{Definition}

\DeclareMathOperator*{\argmax}{arg\,max}

\newcommand{\xx}{{\mathbf{x}}}

\newcommand{\yy}{{\mathbf{y}}}

\newcommand{\ttau}{{\tau}}

\newcommand{\cN}{\mathcal{N}}

\newcommand{\cX}{\mathcal{X}}
\newcommand{\cY}{\mathcal{Y}}

\newcommand{\mI}{\mathsf{I}}

\newcommand{\RR}{\mathbb{R}}

\newcommand{\indicator}{{\mathds{1}}}
\newcommand{\nbhd}{{\mathds{N}}}
\newcommand{\Sset}{{\mathcal{S}}}
\newcommand{\expect}{{\mathds{E}}}

\newcommand{\figref}[2][]{{Figure~\ref{#2}#1}}

\newcommand{\secref}[1]{{Section~\ref{#1}}}

\newcommand{\acro}[1]{\textsc{\MakeLowercase{#1}}}

\newcommand{\mat}[1]{\bm{\mathrm{#1}}}

\renewcommand{\epsilon}{\varepsilon}

\DeclareMathAlphabet{\mathcal}{OMS}{cmsy}{m}{n}  

\begin{document}

%
%

\pagestyle{fancyplain}

\thispagestyle{plain}
\firstPageHead{}

\chead{\fancyplain{}{\itshape Lee, Cheng, and McCourt}}

\rhead{}
\cfoot{}
\renewcommand{\headrulewidth}{0pt} 

\makeatletter
\let\@internalcite\cite
\def\cite{\def\@citeseppen{-1000}%
    \def\@cite##1##2{(##1\if@tempswa , ##2\fi)}%
    \def\citeauthoryear##1##2##3{##1 ##3}\@internalcite}
\def\citeNP{\def\@citeseppen{-1000}%
    \def\@cite##1##2{##1\if@tempswa , ##2\fi}%
    \def\citeauthoryear##1##2##3{##1 ##3}\@internalcite}
\def\citeN{\def\@citeseppen{-1000}%
    \def\@cite##1##2{##1\if@tempswa, ##2)\else{}\fi}%
    \def\citeauthoryear##1##2##3{##1 (##3)}\@citedata}
\def\citeA{\def\@citeseppen{-1000}%
    \def\@cite##1##2{(##1\if@tempswa , ##2\fi)}%
    \def\citeauthoryear##1##2##3{##1}\@internalcite}
\def\citeANP{\def\@citeseppen{-1000}%
    \def\@cite##1##2{##1\if@tempswa , ##2\fi}%
    \def\citeauthoryear##1##2##3{##1}\@internalcite}
\def\shortcite{\def\@citeseppen{-1000}%
    \def\@cite##1##2{(##1\if@tempswa , ##2\fi)}%
    \def\citeauthoryear##1##2##3{##2 ##3}\@internalcite}
\def\shortciteNP{\def\@citeseppen{-1000}%
    \def\@cite##1##2{##1\if@tempswa , ##2\fi}%
    \def\citeauthoryear##1##2##3{##2 ##3}\@internalcite}
\def\shortciteN{\def\@citeseppen{-1000}%
    \def\@cite##1##2{##1\if@tempswa, ##2\else{}\fi}%
    \def\citeauthoryear##1##2##3{##2 (##3)}\@citedata}
\def\shortciteA{\def\@citeseppen{-1000}%
    \def\@cite##1##2{(##1\if@tempswa , ##2\fi)}%
    \def\citeauthoryear##1##2##3{##2}\@internalcite}
\def\shortciteANP{\def\@citeseppen{-1000}%
    \def\@cite##1##2{##1\if@tempswa , ##2\fi}%
    \def\citeauthoryear##1##2##3{##2}\@internalcite}
\def\citeyear{\def\@citeseppen{-1000}%
    \def\@cite##1##2{(##1\if@tempswa , ##2\fi)}%
    \def\citeauthoryear##1##2##3{##3}\@citedata}
\def\citeyearNP{\def\@citeseppen{-1000}%
    \def\@cite##1##2{##1\if@tempswa , ##2\fi}%
    \def\citeauthoryear##1##2##3{##3}\@citedata}
%
%
%
\def\@citedata{%
    \@ifnextchar [{\@tempswatrue\@citedatax}%
                  {\@tempswafalse\@citedatax[]}%
}

\def\@citedatax[#1]#2{%
\if@filesw\immediate\write\@auxout{\string\citation{#2}}\fi%
  \def\@citea{}\@cite{\@for\@citeb:=#2\do%
    {\@citea\def\@citea{, }\@ifundefined
       {b@\@citeb}{{\bf ?}%
       \@warning{Citation `\@citeb' on page \thepage \space undefined}}%
{\csname b@\@citeb\endcsname}}}{#1}}%

%
\def\@citex[#1]#2{%
\if@filesw\immediate\write\@auxout{\string\citation{#2}}\fi%
  \def\@citea{}\@cite{\@for\@citeb:=#2\do%
    {\@citea\def\@citea{; }\@ifundefined
       {b@\@citeb}{{\bf ?}%
       \@warning{Citation `\@citeb' on page \thepage \space undefined}}%
{\csname b@\@citeb\endcsname}}}{#1}}%

%
\def\@biblabel#1{}
\makeatother



\newdimen\bibindent
\bibindent=0.0em
\def\thebibliography#1{\section*{\refname}\list
   {}{\settowidth\labelwidth{[#1]}
   \leftmargin\parindent
   \itemindent -\parindent
   \listparindent \itemindent
   \itemsep 0pt
   \parsep 0pt}
   \def\newblock{}
   \sloppy
   \sfcode`\.=1000\relax}


\setlength{\baselineskip}{12.7pt}

\title{ACHIEVING DIVERSITY IN OBJECTIVE SPACE FOR SAMPLE-EFFICIENT SEARCH OF MULTIOBJECTIVE OPTIMIZATION PROBLEMS }

\author{Eric Hans Lee \\
Bolong Cheng \\
Michael McCourt \\[12pt]
SigOpt, an Intel Company \\
San Francisco, CA, USA \\
}

\maketitle

\section*{ABSTRACT}
Efficiently solving multi-objective optimization problems for simulation optimization of important scientific and engineering applications such as materials design is becoming an increasingly important research topic. This is due largely to the expensive costs associated with said applications, and the resulting need for sample-efficient, multiobjective optimization methods that efficiently explore the Pareto frontier to expose a promising set of design solutions. 
We propose moving away from using explicit optimization to identify the Pareto frontier and instead suggest searching for a diverse set of outcomes that satisfy user-specified performance criteria. This method presents decision makers with a robust pool of promising design decisions and helps them better understand the space of good solutions.  
To achieve this outcome, we introduce the Likelihood of Metric Satisfaction (LMS) acquisition function, analyze its behavior and properties, and demonstrate its viability on various problems.


\section{INTRODUCTION}
\label{sec:intro}

Simulation is a fundamental element to many product and system development processes.  As mathematical, statistical, and machine learning algorithms leverage increasingly powerful computational hardware to perform elaborate tasks, simulation has grown to play a key role in fields such as materials science, operations research, industrial engineering, aerodynamics, pharmaceuticals, image processing, and many others.  In particular, a key use of these simulations is to serve as a surrogate for the eventual implementation and/or manufacturing during the design optimization; running a computational simulation is likely much cheaper than actually conducting a physical experiment or fabrication  \shortcite{forrester2008engineering,negoescu2011knowledge,molesky2018inverse,haghanifar2020}.

Computational simulations can, however, easily run for hours or days, making simulation itself an often costly proposition. The high cost of a single simulation is compounded by the frequent need to simulate many different systems to search for a set of desirable outcomes. This is the motivating force behind simulation optimization, which seeks to identify suitable system parameters to achieve a satisfactory system or effective simulation in a \emph{sample-efficient} fashion, i.e., with as few simulations conducted as possible.  

In practical situations, simulations almost always have multiple competing objectives which define success, and thus it is important for users to understand trade-offs between these competing objectives in order to make an informed design decision. Multiobjective optimization tackles this problem by identifying the \textit{Pareto frontier}, which is the manifold in objective space such that improving one objective cannot occur without harming another. 
Unfortunately, using the Pareto frontier as the measurement of success may be limiting in engineering and design applications. 
Simulations always yield some gap from reality (both due to imprecision in the simulation and  errors in the modeling itself) meaning that even if we could perfectly optimize the simulation, the eventual performance of the real, physical system would still suffer from a loss in optimality.
This ubiquitous problem in single-objective optimization is more pronounced in the multiobjective setting. The Pareto frontier is sensitive to these ever-present inaccuracies, and as a result, candidates near the Pareto frontier, unaccounted for in most optimization literature, are still of scientific interest in the real world.  For example, \shortcite{delrosario2020frontier} introduces the notion of \emph{Pareto shells}, the set of near Pareto optimal solutions, as the desired outcome of multiobjective optimization. 

This sensitivity to both simulation noise and model error make the Pareto frontier inadequate for many practitioners, who seek a more robust method for efficient design-of-experiments. The purpose of this paper is to help bridge this discrepancy between theory and practice by offering an alternative formulation to multiobjective optimization, known as \textit{constraint active search} (CAS) \shortcite{malkomes2021cas}. The purpose of CAS is to account for the three factors described above: sample-efficiency, multiple objectives, and simulation inaccuracy. However, instead of explicit optimization, CAS searches for a diverse set of satisfactory points which can be considered for eventual manufacturing. In \shortcite{malkomes2021cas}, the authors introduced CAS and tackled diversity in parameter space with the expected coverage improvement (ECI) acquisition function. CAS has been successfully used to help design nanostructured glass with multiple desirable optical and physical characteristics, in which it sought a diverse set of candidates in parameter space that fulfilled user-specified performance thresholds. 

In this article, we propose an alternate definition of diversity to guide constraint active search.  Rather than using diversity in parameter space (the input space), we seek diversity in \textit{objective space} (the output space).  Doing so gives us the ability to have greater understanding as to how the competing objectives are likely to interact among satisfactory configurations; this, in turn, gives us more confidence about the eventual manufacturing or deployment process.
Our paper makes the following concrete contributions:

\begin{itemize}
    \item We propose an alternative approach to multiobjective optimization, in which we solicit minimum performance thresholds and search for a diverse set of objective values that meet these thresholds. 
    \item We use multi-output Gaussian process modeling combined with a novel acquisition function called Likelihood of Metric Satisfaction (LMS) to search for a diverse set of feasible objective values in a sample-efficient manner. 
    \item We demonstrate the effectiveness on a set of synthetic problems as well as a real-world nuclear fusion simulation optimization problem.
    \item We identify key practical steps to take when using CAS-LMS to achieve diversity in objective space in adverse and/or complicated circumstances.
\end{itemize}

\section{BACKGROUND}
\label{sec:background}

\subsection{Bayesian Optimization}
One of the best-known stochastic surrogate optimization methods for expensive, black functions is Bayesian optimization (BO). BO consists of two core components: a probabilistic model (commonly referred to as a \emph{surrogate model}) that models the objective function and an \emph{acquisition function} that uses the model to determine where next to sample.

BO research can be traced back to the Efficient Global Optimization (EGO) algorithm \shortcite{jones1998ego}, which combined a Gaussian Process model with the Expected Improvement acquisition function \shortcite{Mockus1978ei}. Recent research has proposed the use of other probabilistic models, such as kernel density estimators \shortcite{bergstra2011tpe} or random forests \shortcite{hutter2011smbo}. In practice, Gaussian processes \shortcite{Rasmussen_gpforml} have become the predominant surrogate model of choice in the BO community, bolstered by the proliferation of modern GP software such as GPy \shortcite{gpy2014} and GpyTorch \shortcite{gardner2018gpytorch}. 
A significant portion of the BO literature focuses on proposing new acquisition functions, including Upper Confidence Bound \shortcite{srinivas2010gpucb}, Knowledge Gradient \shortcite{scott2011gpkg}, and Entropy Search \shortcite{villemonteix2009information}. Each of these methods have different treatment of the exploration exploitation tradeoff. As a result, there is no one default acquisition function agreed upon in practice, and researchers continue to develop new variations of these aforementioned acquisition functions for different applications. We refer the readers to \shortcite{garnett_bayesoptbook_2022} for a modern comprehensive review of BO.

\subsection{Multiobjective Bayesian Optimization}

The earliest attempt at adapting BO to multiobjective optimization is the Pareto Efficient Global Optimization (ParEGO) algorithm \shortcite{knowles2006parego}, which uses 
the aforementioned EGO algorithm to optimize a linear combination of the multiple objectives. This so-called linear scalarization does not work well when the Pareto frontier is non-convex in the objective space. Second, this method lacks interpretability in practice; finding the appropriate weighting of multiple objectives is a nontrivial task.

A separate approach leverages work from constrained Bayesian optimization literature \shortcite{gardner_boconstraints}, by reformulating the problem as a constrained optimization problem, commonly known as the $\epsilon$-constraint method.
In this setting, the objectives are treated as inequality constraints of the form $f(\xx) \leq \epsilon$, for some known threshold $\epsilon$. These constraints often appear naturally in real world applications, such as baseline performance metrics \shortcite{haghanifar2020}.

Recent effort in multiobjective BO focuses on directly improving the hypervolume of the Pareto frontier directly \shortcite{emmerich2011hvei,daulton2020ehvi,daulton2021multi}. In particular, \shortcite{daulton2020ehvi} exploits modern parallel hardware (such as GPUs) to efficiently compute the expected hypervolume. Similar to $\epsilon$-constraint methods, these hypervolume improvement methods also require knowing \emph{a priori} thresholds on the objective values.

\subsection{Active Search and Feasibility Determination}
Our proposed work is also related to the topics of \textit{active search} \shortcite{garnett2012active,jiang2017enas} and \textit{feasibility determination} \shortcite{szechtman2008new,szechtman2016bayesian}. 
Active search can be seen as a special case of Bayesian optimization, where one has binary observations and cumulative reward. 
The goal of active search is to sequentially discover members of a rare, desired class. 
Inspecting any element is assumed to be expensive, representing, for instance, the cost of performing a real world laboratory experiment.


\subsection{Constraint Active Search with Parameter Diversity}
\label{sec:cas}
Constraint active search is an alternate formulation of the multiobjective design problem first proposed in \shortcite{malkomes2021cas}.  In lieu of identifying the highest performing design (in a single objective setting) or the Pareto frontier (in a multiobjective setting), success is defined as the identification of multiple outcomes in a ``satisfactory region''. User-defined upper or lower constraints on each objective implicitly define the satisfactory region (the region of satisfactory performance); we adopt the same structure. Designs which satisfy these constraints are termed satisfactory.

CAS proceeds in a sequential fashion, similarly to Bayesian optimization: first, each objective is modeled using the available data (we use a Gaussian process, though other models are viable); second, a next design at which to sample is suggested through optimization of an acquisition function.  \shortcite{malkomes2021cas} introduced the expected coverage improvement (ECI) acquisition function -- ECI is maximized for designs which have high probability of satisfying the user-defined constraints but also are sufficiently distinct from previously observed satisfactory designs in parameter space.

\section{CONSTRAINT ACTIVE SEARCH WITH OBJECTIVE VALUE DIVERSITY\label{sec:metric-def-header}}
\begin{figure}[t] 
    \centering
    \includegraphics[width=\textwidth]{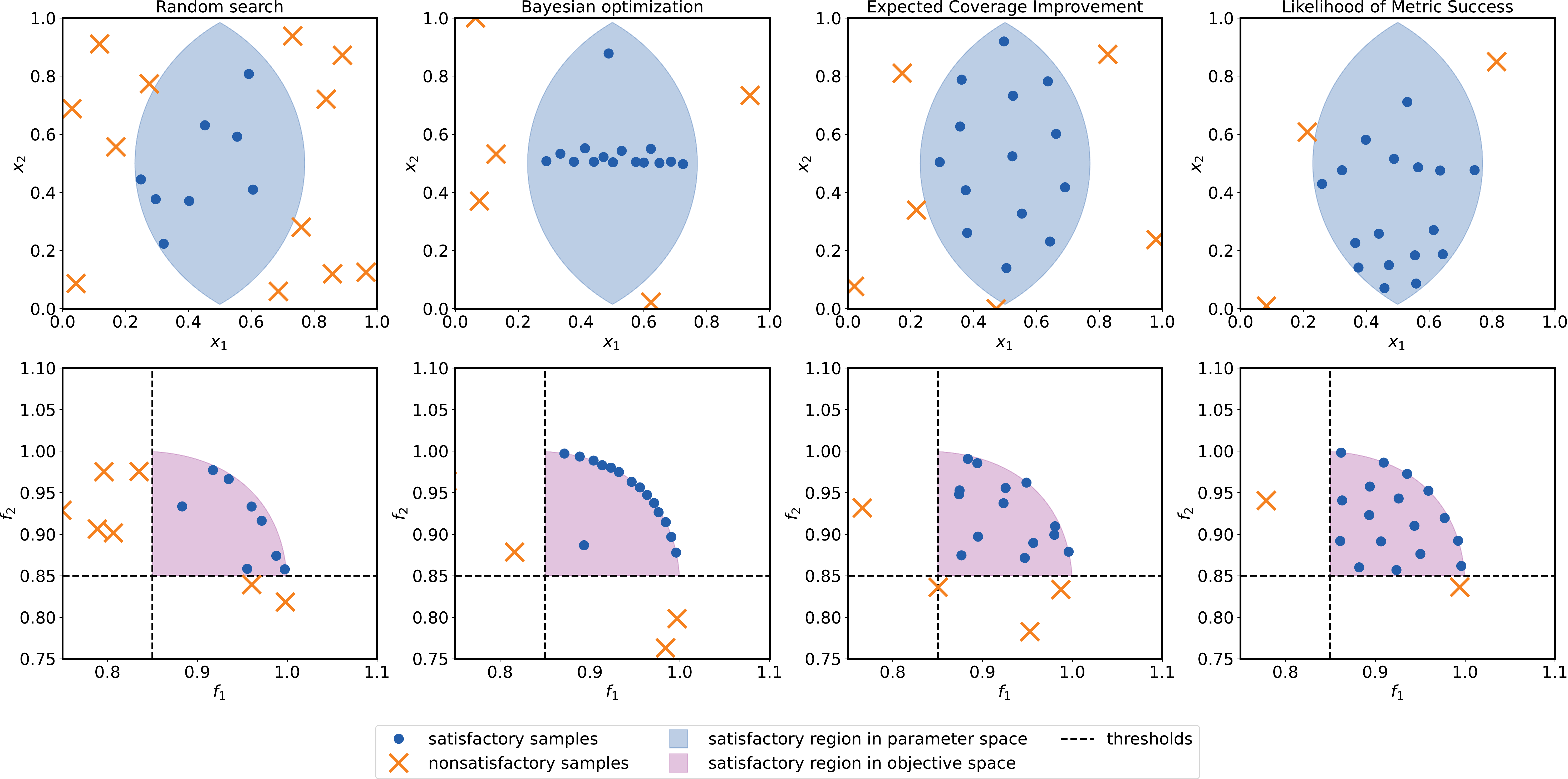}
    \caption{We compare random search, BO, CAS in parameter space, and CAS in objective space on the same multiobjective problem. 
    The top row shows the observed samples in the parameter space (blue); the bottom row, in the objective space (red). BO seeks to isolate the Pareto frontier, leading to a tight concentration of points in both parameter and objective space. CAS in parameter space unsurprisingly achieves diversity in parameter space, but at the cost of poor spread in objective space. Spreading points out in objective space ---CAS with objective value diversity--- is the topic of this paper. 
}
    \label{fig:search_comparison}
\end{figure}

Given a target range of objective values, we seek to generate a diverse set of outcomes in the objective space that presents decision makers with a robust set of promising designs, so that they may better understand the space of good solutions for the problem at hand. We want to do this in a sample-efficient manner, which presents two additional challenges that are not present in the parameter space formulation of constraint active search. 

\begin{itemize}
    \item \textbf{Objective Uncertainty:} Achieving diversity in objective space is subject to uncertainty because we do not know prior to evaluating the objective what those values will be. This stands in contrast to achieving parameter diversity, in which no uncertainty exists in the parameter values. 
    \item \textbf{Objective Heteroskedasticity:} Sampling the objective uniformly across parameter space will very likely yield a non-uniform distribution of objective values. For example, a substantial portion of the domain in parameter space may map to a very concentrated cluster in objective space, making it difficult for an algorithm to spread objective values out efficiently.  
\end{itemize}

To address these challenges, we propose a novel acquisition function criterion ---which we call Likelihood of Metric Satisfaction (LMS)--- to select a suitable next evaluation during constraint active search. LMS is defined over our optimization domain $\cX$ and the value $\text{LMS}(\xx)$ quantifies how much $\xx$ might improve the diversity in objective values we seek. 

We outline our constraint active search procedure with the following general steps:
\begin{enumerate}
    \item Build a multi-output Gaussian process surrogate to model the observation data. 
    \item Choose the next evaluation to be $\xx^* = \argmax_\cX \text{LMS}(\xx)$. 
    \item Evaluate $\yy = f(\xx)$, update the observation data, and repeat until budget exhausted. 
\end{enumerate}

In the following subsections, we lay out a more precise problem statement, describe the Gaussian process (GP) model, and then explain LMS at length. 

\subsection{Notation and Problem Statement }
Suppose we want to search for design configurations in a $d$-dimensional, compact search space $\cX \subset \RR^d$. 
We may judge the quality of a design $\xx \in \cX$ by evaluating $m$ expensive black-box objective functions $f_1, f_2, \ldots f_m$, each mapping $\cX$ to $\RR$. 
We seek designs $\xx$ that yield satisfactory performance, 
defined by finite threshold values $\ttau = [\tau_1, \tau_2, \ldots, \tau_m]^\top$.
Specifically, we wish to sequentially select configurations from:
\[
\Sset = \{ \xx \;|\; f(\xx) \succeq \ttau \},
\]
where $f(\xx) \succeq \ttau := f_i(\xx) \geq \tau_i, \; i=1, \ldots, m$. We refer to $\Sset$ as the \emph{satisfactory region}. We make the assumption that $f: \cX \rightarrow \cY$ is continuous and that the set $\{ \yy \;|\; \yy \succeq \tau\}$ is also compact. 

The concrete goal of CAS using the LMS acquisition function is identify candidates in $\Sset$ and disperse them throughout the subregion of objective space that exceeds a certain user-defined performance threshold. The LMS acquisition function does this by attempting to guarantee that each point is at least distance $r$ from any other point in objective space, where $r$ is a user-defined resolution parameter. 


\subsection{Gaussian Process Models\label{sec:gaussian-process}}

A Gaussian process (GP) model is a specific type of stochastic surrogate model ---to be more precise, a random field--- that uses radial basis functions to model the value of an unseen point as a normal distribution \shortcite{FasshauerGE2015ws}. In recent years, GPs have become a popular method for modeling simulation experiments \shortcite{binois2015quantifying,binois2018practical}, due to not only their ability to accurately approximate a wide range of continuous functions, but also due to their built-in uncertainty estimates. 

We start by first describing the GP model of a scalar function. Note that for clarity, the notation used in this subsection will be slightly different from that of the rest of the paper. We assume that we are trying to model a function $y = f(\xx)$, $\xx \in \RR^d$, $\yy \in \RR$, and $f(\xx): \RR^d \rightarrow \RR$. We collect $n$ observations of $f(\xx)$ in the pair of matrices $\{\mat{X}, \yy \}$, where $\mat{X} = [\xx_1, \dots, \xx_n]$ and $\yy = [y_1, \dots y_n]^T$. 

On this set of $n$ observations we place a GP prior. Given this GP prior, the posterior distribution of any set $k$ of function values is modeled by a random variable $y_\xx \in \RR^k$ with the following normal distribution:

\[
y_\xx \sim \cN ( \tilde \mu_\xx, \tilde K_{\mat{X} \mat{X}}) \;|\;
\tilde \mu_\xx = K_{\mat{X} \xx}^T (K_{\mat{X} \mat{X}} + \sigma^2 \mI_d)^{-1}(\yy_\xx - \mu_\xx) + \mu_\xx \;,\;
\tilde K_{\mat{X} \mat{X}} = K_{ \mat{X} \xx}^T(K_{\mat{X} \mat{X}} + \sigma^2 \mI_d)^{-1}K_{\mat{X} \xx }.
\]

The entries of the vector $\mu_\xx \in \RR^n$ and the matrices $K_{ \mat{X}x } \in \RR^{n \times k}$, $K_{\mat{X} \mat{X}} \in \RR^{n \times n}$ are determined by a mean function $\mu(\xx)$ and a covariance kernel $k(\xx, \xx')$ respectively, which largely control the fit of the GP. The $\sigma^2$ term is a regularization parameter and $I$ is the $n \times n$ identity matrix. The GP we have described models a scalar function. However, in this paper we deal with multi-objective functions. To model these, we simply use a collection of $m$ independent GP models. 


\subsection{Constraint Active Search with Parameter Diversity\label{sec:parameter-diversity}}
Constraint active search (CAS) was first introduced in \shortcite{malkomes2021cas}, in which the authors searched for designs in $\Sset$ which were as dispersed as possible within $\cX$.  The ECI acquisition function was created in pursuit of this goal. ECI balances the exploitation and exploration tradeoffs by simultaneously preferring candidates that are a) likely to satisfy the performance thresholds and b) located in unexplored regions. This paper can be viewed as an extension of CAS to objective space instead of the parameter space considered in the original work. The challenges of achieving diversity in parameter and objective space are somewhat different given the heteroskedasticity and uncertainty in the latter, and therefore our approach in this paper differs from that of ECI. However, we maintain a resolution parameter $r$ which defines a sense of locality through Euclidean distance surrounding satisfactory designs.


\subsection{Likelihood of Metric Satisfaction}
\label{sec:metric diversity}

\begin{figure}[t]
    \centering
    \includegraphics[width=\textwidth]{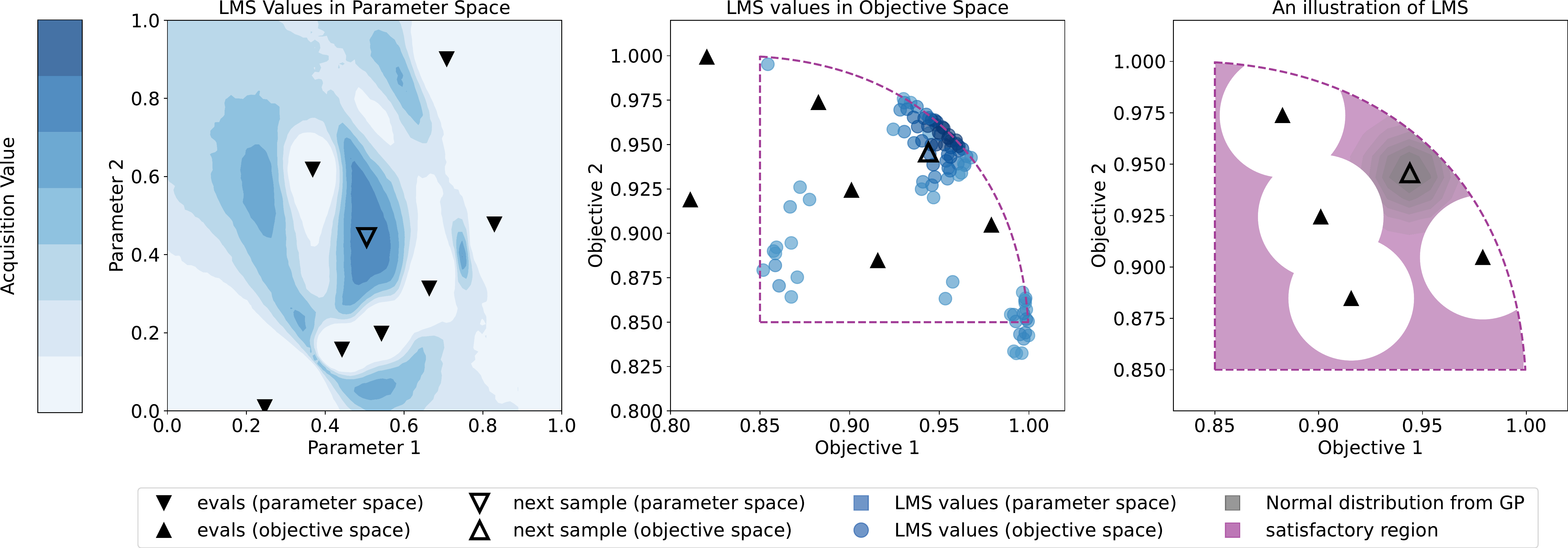}
    \caption{ On the left, we plot the LMS values in parameter space. In the middle, we plot high values of LMS in objective space. On the right, we visualize computing LMS($\xx^*$), where $\xx^*$ is. LMS over our domain, in objective space. The LMS value is the volume of the Gaussian probability distribution that models $f(\xx^*)$ centered around $\mu(\xx^*)$ that is within the satisfactory region (magenta) and sufficiently far from existing observations in objective space (white circles). }
    \label{fig:emc_illustration}
\end{figure}

LMS quantifies how likely a configuration $\xx$ will be satisfactory (exceeding all thresholds) and promote diversity in objective space. This requires us to first define what diversity means. For LMS, we specify diversity using an additional parameter $r$, which informs the minimal distance between objective values it seeks to achieve. Given $r$, we define the following. 
\begin{definition}[Number of neighbors within $r$]
\textit{The number of neighbors within $r$ of a point $\yy$ and a finite set $\mat{Y} \subset \mathcal{Y}$ is defined as
\[
\nbhd_{r}(\yy, \mat{Y}) = |\{ \yy'  \;|\; \yy' \in \mat{Y}, d(\yy, \yy') < r \}| ,
\]
for an a priori fixed $r \in \RR^+$ and an appropriate distance function $d : \cY \times \cY \mapsto \RR^+$. }
\end{definition}

\begin{definition}[Average number of neighbors within $r$]
\textit{The average number of neighbors within $r$ of a finite set $\mat{Y} \subset \mathcal{Y}$ is defined as
\[
\nbhd_{r}(\mat{Y}) =  \frac{1}{n} \sum_{i = 1, \yy_i \in \mat{Y}}^n \nbhd_r(\yy_i, \mat{Y}),
\]
for an a priori fixed $r \in \RR^+$ and an appropriate distance function $d : \cY \times \cY \mapsto \RR^+$. }
\end{definition}
\begin{definition}[Comparing the spatial diversity of two sets]
\textit{Given a radius $r$, we say that $\mat{Y}_1$ exhibits more spatial diversity than $\mat{Y}_2$ if 
$\nbhd_{r}(\mat{Y}_1)  < \nbhd_{r}(\mat{Y}_2).$}
\end{definition}
In other words, $\mat{Y}_1$ exhibits greater spatial diversity than $\mat{Y}_2$ if it has fewer neighbors within $r$ on average. More generally, a set of points that are well-spaced apart will be more spatially diverse than a set of points that are very tightly clustered. Note that this definition is somewhat counter-intuitive because a lower $\nbhd_{r}$ implies higher diversity.

We need one last definition and that is the half space whose points are greater than the threshold. 

\begin{definition}[Half space exceeding thresholds]
\textit{$H_\tau \subset \RR^m$ is the half-space greater than thresholds $\tau$:
\[
H_\tau = \{ \yy \;|\; \yy \in \RR^m, \yy_i \geq \tau_i, i=1, \dots\ m \}.
\]}
\end{definition}

Having now defined all these things, we can now precisely define LMS. Assume we already have a set of in observations in objective space $\mat{Y}$. We want the objective values of the next observation $\yy$ to decrease $\nbhd_{r}(\mat{Y} \cup \{\yy \})$, which is equivalent to increasing diversity in objective space. 
This is guaranteed to occur if $\nbhd_{r}(\yy, \mat{Y}) = 0$; $\yy$ is not within $r$ of any other observation. 
We also want $\yy$ to satisfy our thresholds i.e., $\yy \in H_\tau$. LMS is the probability of these two events occurring with respect to a GP distribution on $\yy$.
\begin{definition}[Likelihood of Metric Satisfaction]
\textit{The LMS value of a point $\xx$ and it's associated objective values $\yy = f(\xx)$ is the probability that $\yy$ has no neighbors within $r$ and lies in $H_\tau$:}
\[
\text{LMS}(\xx) = Pr\bigg( \nbhd_{r}(\yy, \mat{Y}) = 0  \;, \; \yy \in H_\tau \bigg).
\]
\end{definition}
Thus, the higher the LMS value, the greater the probability that the observed next objective value $\yy$ will decrease $\nbhd_{r}$; thus, it may help to think of LMS as attempting to minimize $\nbhd_{r}$ in a sample-efficient manner. 

We assume we have $m$ independent Gaussian process models that capture our prior beliefs about observations $y_i = f_i(\xx_i) + \epsilon_i$ for $i=1, 2, \ldots, m$, as a probability distribution over $p(\yy)$, where $\epsilon_i$ is additive Gaussian noise. This set of GPs models each point $\yy(\xx) = [y_1(\xx) \dots y_m(\xx)]$ as the following distribution:
\[
y_i(\xx) \sim \cN( \tilde \mu^{(m)}_\xx , \tilde \Sigma^{(m)}_{\xx, \xx}) \;|\; \tilde \mu^{(m)}_\xx \in \RR^m, \tilde \Sigma^{(m)}_{\xx, \xx} \in \RR^{m \times m}, 
\]
where $\tilde \mu^{(m)}_\xx$ is vector of GP means for each objective and $\tilde \Sigma^{(m)}$ is the matrix of GP variances for each objective. 
Then LMS($\xx$) is the the volume of the PDF of $\cN( \tilde \mu^{(m)}_\xx , \tilde \Sigma^{(m)}_{\xx, \xx})$ that is above the thresholds and not within radius $r$ of any existing observation in objective space. This is visualized in Figure \ref{fig:emc_illustration}, in which we plot the LMS values over the parameter and objective spaces.

To compute LMS, we rewrite it as the following integral of an indicator function: 
\[
\text{LMS}(\xx) = \expect_\yy[ \indicator_{\tau, \mat{Y}}(\yy_\xx)] = \int_{\RR^m} \indicator_{\tau, \mat{Y}}(\yy_\xx) p(\yy_\xx) d\xx,
\]
where $\indicator_{\tau, \mat{Y}}(\yy_\xx)$ is 1 when $\yy$ is above the thresholds and outside $r$ of any observations, and 0 otherwise. 
We compute LMS in a straightforward manner with Monte Carlo (MC) integration ---sample $\{\yy_1, \dots, \yy_ N\}$ from $\cN( \tilde \mu^{(m)}_\xx , \tilde \Sigma^{(m)}_{\xx, \xx})$ and sum the samples:
$
\text{LMS}(\xx) \approx \frac{1}{N} \sum_{i=1}^N \indicator_{\tau, \mat{Y}}(\yy_i) .
$
We note that computing the acquisition function via MC is very common in Bayesian optimization \shortcite{emmerich2011hvei,daulton2020ehvi,garnett_bayesoptbook_2022} and is indeed supported by popular BO packages such as BoTorch \shortcite{balandat2020botorch}. 

\subsection{Scaling in Objective Space}
When ECI is used for diversity in parameter space, the user benefits from knowing \emph{a priori} what range the parameter values may take ---indeed, they are the ones who define the search domain $\cX$. As a result, ECI is able to choose an appropriate measure of distance $d(\xx, \xx')$ such that scale of each axis is the same ---for example, by normalizing $\cX$ to be the unit hypercube $[0, 1]^d$. 

No such luxury exists in the case of LMS, which works with no prior knowledge of the range of $\cY$. If objective $f_1$ represents simulation time and objective $f_2$ represents a residual, then $f_1$ may be on the order of $10^{6}$  while $f_2$ may be on the order of $10^{-6}$. In this particular case, if the user selects $d(\yy, \yy')$ to be standard Euclidean distance, any perturbation to $f_1$ will far exceed the largest of perturbations to $f_2$; this heavily biases diversity towards $f_1$. 

There is no perfect solution to this problem, and we advocate for dynamic scaling of each objective's axis to address this problem. At each iteration, we consider the minimal bounding box of our observed objective values and scale each dimension to be unit length. We provide an experimental study in \secref{sec:experiments_scaling} to illustrate the practical improvements that dynamic scaling provides. 

\section{EXPERIMENTS}
\label{sec:experiments}
In this section, we present numerical experiments to analyze the efficacy of our method. We compare our LMS acquisition function against three baselines: Random search, expected hypervolume improvement \shortcite{daulton2020ehvi}, and the expected coverage improvement \shortcite{malkomes2021cas}.

We emphasize that no single criterion can sufficiently convey the full strength of any methodology; this is especially true in the multiobjective setting, in which different performance criteria already exist to quantify different algorithmic goals. We hope to impart a nuanced comparison of LMS to existing baselines. 
To that end, we consider the following four standard criteria. 

\begin{itemize}
\item  \textbf{Number of neighbors within $r$}: Defined in Section \ref{sec:metric diversity}, LMS attempts to keep this quantity small. Note that this is a particular metric for the more general notion of coverage, which is a commonly used criteria to judge the spread of points in a continuous space \shortcite{sayin2000measuring}

\item \textbf{Fill distance}: Fill distance is a standard measure of spatial diversity in a simple domain $\Sset$. Given a set of sample points $\mat{Y}$, the fill distance is formally defined as the following:
$
 \acro{fill}(\mat{Y}, \Sset) =  \sup_{\yy \in \Sset} \min_{\yy_j \in \mat{Y}} d(\yy_j , \yy). 
$
In Euclidean space, $\acro{fill}(\mat{Y}, \Sset)$ is the radius of the largest empty ball one can fit in $\Sset$, and measures the spacing of $\mat{X}$ in $\Sset$. The smaller a set's fill, the better distributed it is within $\Sset$. Special sets that achieve low fill in simple domains include low-discrepancy sequences and Latin hypercubes, used frequently in simulation optimization \shortcite{niederreiter1992random}. 
\item \textbf{Positive samples}: The number of points whose objective values exceed $\ttau$. 
\item \textbf{Hypervolume}: We measure the hypervolume of region in objective space bounded by the Pareto frontier and the defined thresholds. 
In particular, we conjecture that algorithms which excel at maximizing the hypervolume may underperform on other criteria, and vice versa.
\end{itemize}

\begin{table*}[t]
  \centering
\caption{
  Select experimental results from three problems: HC22, described in \secref{sec:base-exp}, RE33 \protect\shortcite{tanabe2020reproblems}, and a plasma physics simulation optimization problem, with an optimization budget of 30, 50, and 100, respectively. The median over ten trials is provded below. We see that LMS achieves the lowest fill distance in objective space. It also generally achieves the lowest number of neighbors within $r_\yy$ (random search does this trivially by identifying very few feasible points, thus guaranteeing that they are spaced far apart). MOBO achieves the highest hypervolume. 
}
\begin{tabular}{@{}llllcccc@{}}
\toprule
Function               & $d$                   & $m$                                   & Methods & Fill distance $\downarrow$ & \# Satisfactory $\uparrow$ & \# Neighbor $\downarrow$ & Hypervolume $\uparrow$ \\ 
\midrule
\multirow{4}{*}{HC22} & \multirow{4}{*}{2} & \multirow{4}{*}{2}  & \acro{RND} & 4.31 & 2 & \textbf{0.0} & 0.16 \\
& & &  \acro{BO} & 11.4 & \textbf{27} & 9.20 & \textbf{1.71} \\
& & &  \acro{ECI}($r_\xx=0.1$) & 4.84 & 20 & 2.5 & 1.50 \\
& & &  \acro{LMS}($r_\yy=0.1$) & \textbf{2.61} & 21 & 1.0 & 1.50 \\
\midrule
\multirow{4}{*}{RE33} & \multirow{4}{*}{3} & \multirow{4}{*}{3}  & \acro{RND} & 7.18 & 5 & \textbf{0.0} & 5.71 \\
& & &  \acro{BO} & 2.53 & 32 & 3.06 & \textbf{15.02} \\
& & &  \acro{ECI} ($r_\xx=0.1$) & 10.01 & \textbf{45} & 4.71 & 10.52 \\
& & &  \acro{LMS} ($r_\yy=0.26$) & \textbf{1.50} & 33 & 2.48 & 6.93 \\
\midrule
\multirow{4}{*}{STELL} & \multirow{4}{*}{9} & \multirow{4}{*}{3}  
& \acro{RND}                    & 37.62 & 7 & \textbf{0.0} & 17.04 \\
& & &  \acro{BO}                & 31.31 & 30 & 15.46 & \textbf{36.83} \\
& & &  \acro{ECI} ($r_\xx=0.1$) & 32.66 & \textbf{32} & 14.88 & 25.99 \\
& & &  \acro{LMS} ($r_\yy=0.5$) & \textbf{30.79} & 20 & \textbf{0.0} & 29.18 \\
\midrule
\bottomrule
\end{tabular}
\end{table*}

\subsection{Demonstration of Likelihood of Metric Satisfaction}
\label{sec:base-exp}

\label{sec:experiments_scaling} 
\begin{figure}[ht!]
    \centering
    \includegraphics[width=0.9\textwidth]{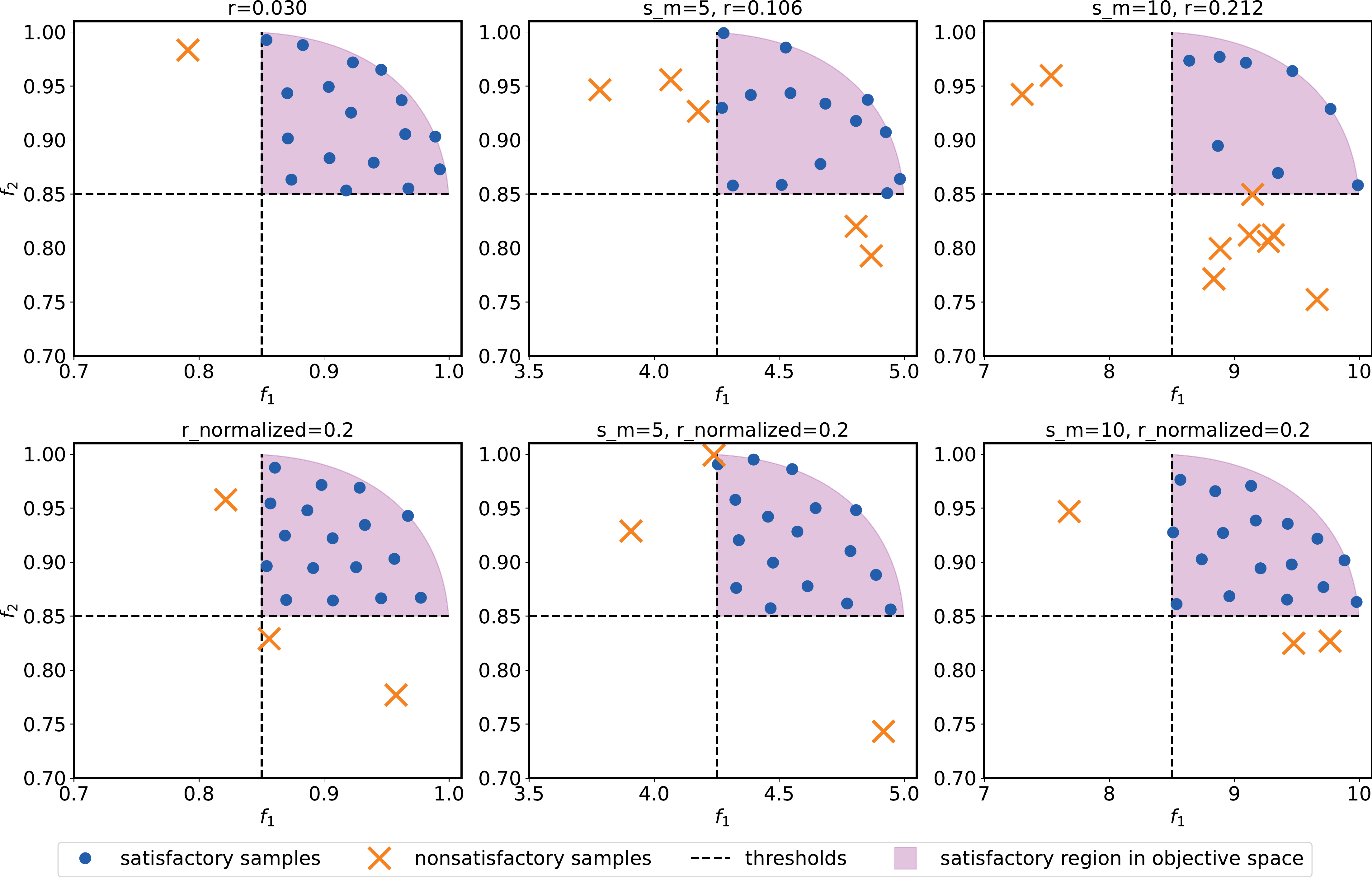}
    \caption{Comparison of non-normalized LMS (top row) and normalized (bottom row) LMS for different scaling of $f_1(\xx)$. We show the samples and satisfactory region in objective space for scaling $s_m = \{5, 10\}$.}
    \label{fig:emc_y_scaling}
\end{figure}

Here we compare the algorithms on a simple two objective problem, with each objective $f: R^2 \rightarrow R$,
\begin{align*}
f_1(\xx) &= \exp(((x_1 - 0.2)^2 + (x_2 - 0.5)^2) / 2),  \\
f_2(\xx) &= \exp(((x_1 - 0.8)^2 + (x_2 - 0.5)^2) / 2).
\end{align*} 
We set the thresholds to be $f_1(\xx) \geq 0.85$ and $f_2(\xx) \geq 0.85$. We run each method for 20 iterations and plot the samples in both parameter space and objective space in \figref{fig:search_comparison} of \secref{sec:cas}. 
Visually, we see that multiobjective Bayesian optimization seeks to isolate the Pareto frontier, leading to a tight concentration of points in both parameter and objective space. CAS using ECI does a good job of spreading points out in parameter space, but not objective space. Conversely, using LMS instead yields our desired result ---a diverse spread of points in objective space.

\subsection{Scaling in Objective Space}

Next, we investigate how the difference in the range of $f$ can impact the LMS algorithm.  We repeat the test functions in \secref{sec:base-exp}, but scale $f_1(\xx)$ and its corresponding threshold by a factor of $s_m = \{5, 10\}$. For the non-normalized LMS, we multiply the resolution parameter $r$ by a factor of $s_m /\sqrt 2$. For the normalized LMS, we fix the $r_{normalized} = 0.2$. We demonstrate the different behaviors of the normalized and non-normalized LMS algorithms in \figref{fig:emc_y_scaling}.

We can observe that the non-normalized LMS algorithm tends to search along $f_1$ when scaled, since it is more likely find new satisfactory samples that are at least $r$ away from the observed samples because the axis along $f_1$ is longer. In contrast, the normalized LMS algorithm consistently produces more ``evenly'' distributed samples in the objective space despite the scaling of $f_1$.

\subsection{Relationship between Parameter and Objective Satisfactory Region}

\begin{figure}[ht!]
    \centering
    \includegraphics[width=\textwidth]{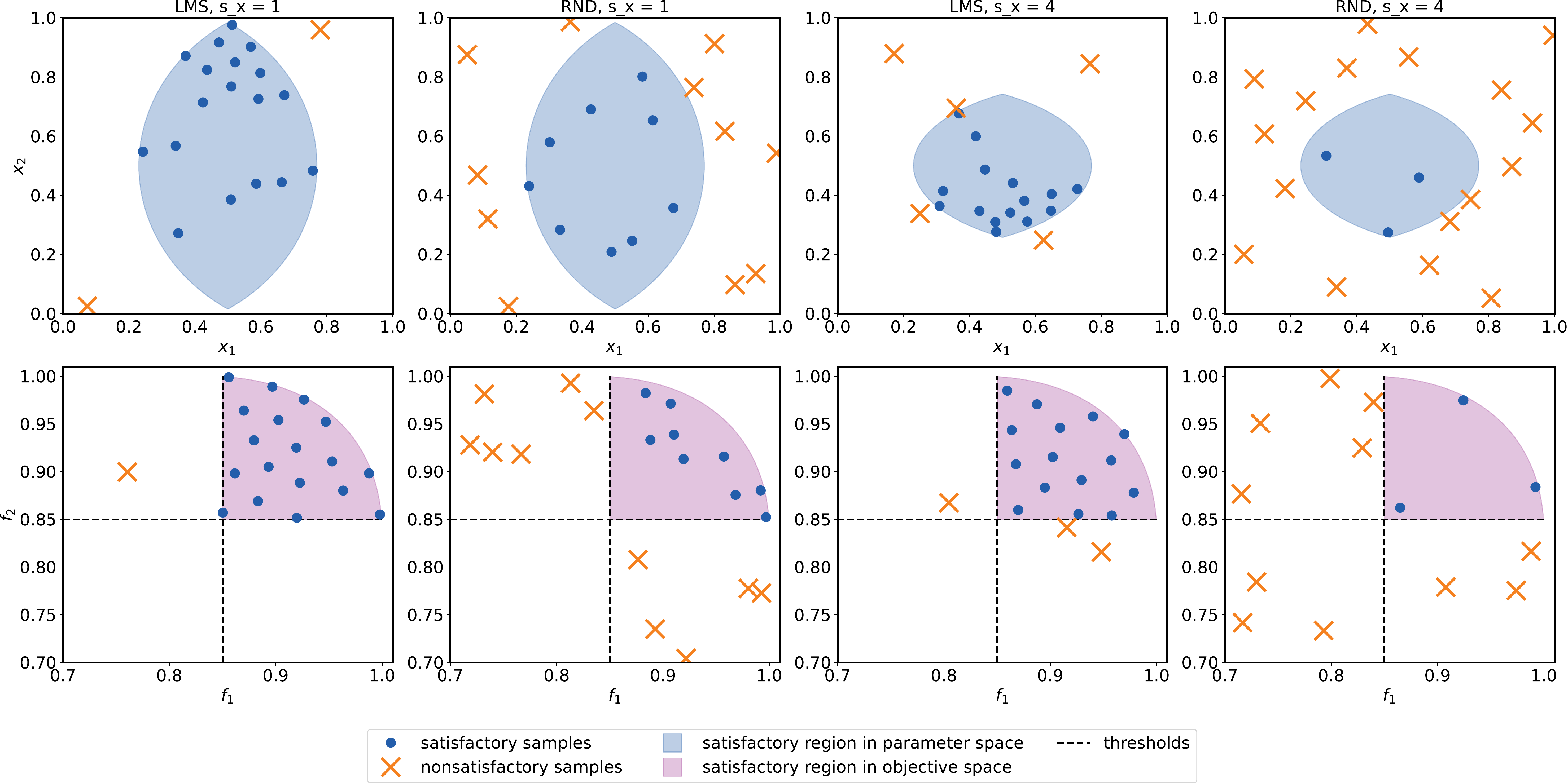}
    \caption{Comparison of LMS and RND for different parameter satisfactory volumes (same satisfactory region in objective space).}
    \label{fig:lms_parameter_metric_satisfied}
\end{figure}


We then investigate how LMS perform for different parameter satisfactory regions. More specifically, we want to see if LMS still works as the relative satisfactory region in the parameter space decreases. We modify the objective functions in \secref{sec:base-exp} with a $s_m$ scaling factor. As $s_m$ increases, the satisfactory region in the parameter space decreases, but the satisfactory region in the objective space remains the same.

\begin{align*}
f_1(\xx) &= \exp(((x_1 - 0.2)^2 + s_m (x_2 - 0.5)^2) / 2),  \\
f_2(\xx) &= \exp(((x_1 - 0.8)^2 + s_m (x_2 - 0.5)^2) / 2).
\end{align*} 

We demonstrate in \figref{fig:lms_parameter_metric_satisfied} that LMS can consistently sample diversely in the satisfactory region in the objective space, despite the satisfactory region in the parameter space shrinking. In contrast, random search is unable to handle the shrinking satisfactory region.

\subsection{Plasma Physics}
A stellarator is device that uses a set of magnetic coils to confine a plasma hot enough to sustain nuclear fusion \shortcite{spitzer1958stellarator}. Stellarator coils lack rotational symmetry, and possess notably warped shapes due to the complex, quasi-symmetric magnetic field they must produce. Thus, determining suitable stellarator designs through simulation is crucial to produce a stellarator candidate for real-world production.

We use the PyPlasmaOpt \shortcite{giuliani2020single} simulator to generate a diverse set of stellarator designs, where each stellarator is a set of coils. Each coil is represented by the curve in 3D Cartesian coordinates $\Gamma(\theta) = (x(\theta), y(\theta), z(\theta))$, where each coordinate admits the following Fourier expansion, e.g., for $x$:
\[
x(\theta) = c_0 + \sum_{k=1}^{n_{order}} s_k\sin(k\theta) + c_k\cos(k\theta).
\]
For each coil, the parameters $c_k$ and $s_k$ are the search parameters. Given fixed coils, the simulator solves a certain first order, nonlinear ordinary differential equation outputs summary information in three functions:
\[
f(\xx) = [F_{\text{magnetic}}(\xx) , F_{\text{transform}}(\xx) , F_{\text{shape}}(\xx)].
\]
The first quantifies the quasi-symmetry of the magnetic field ---the smaller it is, the more desirable the resulting field. The second locks the solution into a target rotational transform. The third penalizes overly complex coils too impractical to manufacture in real life. We test LMS on a small stellarator simulation of six repeated coils and present results in Table 1. 


\section{CONCLUSION}
In this paper, we tackle an alternative formulation multi-objective optimization problems for simulation optimization by generating a diverse set of designs in objective space instead of explicitly searching for the Pareto frontier. This presents decision makers with a robust pool of promising design decisions and helps them better understand the space of good solutions.  

We do so through an acquisition called Likelihood of Metric Satisfaction (LMS), which quantifies the utility of a candidate point as the probability that it is both above user-defined performance thresholds and sufficiently far from other observations in objective space. We then illustrated the strength of LMS on a few synthetic simulation optimization problems as well as one application in stellarator design for nuclear fusion. Finally, we examine the performance of LMS under adverse conditions. 
We believe LMS and more generally, the presentation of multiple, diverse solutions to a design problem, is a promising research direction. Future work includes simultaneously considering diversity in both parameter and objective space. 

\label{sec:conclusion}

\footnotesize

\bibliographystyle{wsc}

\bibliography{references}

\begin{thebibliography}{}

\bibitem[\protect\citeauthoryear{Balandat, Karrer, Jiang, Daulton, Letham,
  Wilson, and Bakshy}{Balandat et~al.}{2020}]{balandat2020botorch}
Balandat, M., B.~Karrer, D.~Jiang, S.~Daulton, B.~Letham, A.~G. Wilson, and
  E.~Bakshy. 2020.
\newblock ``BoTorch: A Framework for Efficient Monte-Carlo Bayesian
  Optimization''.
\newblock In {\em Advances in Neural Information Processing Systems}, edited
  by\ H.~Larochelle, M.~Ranzato, R.~Hadsell, M.~Balcan, and H.~Lin, Volume~33.
\newblock December 6\textsuperscript{th}-12\textsuperscript{th}, virtual,
  21524--21538.

\bibitem[\protect\citeauthoryear{Bergstra, Bardenet, Bengio, and
  K\'{e}gl}{Bergstra et~al.}{2011}]{bergstra2011tpe}
Bergstra, J., R.~Bardenet, Y.~Bengio, and B.~K\'{e}gl. 2011.
\newblock ``Algorithms for Hyper-Parameter Optimization''.
\newblock In {\em Advances in Neural Information Processing Systems}, edited
  by\ J.~Shawe-Taylor, R.~Zemel, P.~Bartlett, F.~Pereira, and K.~Q. Weinberger,
  Volume~24.
\newblock Dec 12\textsuperscript{th}-17\textsuperscript{th}, Grenada, Spain.

\bibitem[\protect\citeauthoryear{Binois, Ginsbourger, and Roustant}{Binois
  et~al.}{2015}]{binois2015quantifying}
Binois, M., D.~Ginsbourger, and O.~Roustant. 2015.
\newblock ``Quantifying Uncertainty on Pareto Fronts with Gaussian Process
  Conditional Simulations''.
\newblock {\em European Journal of Operational Research\/}~243(2):386--394.


\bibitem[\protect\citeauthoryear{Binois, Gramacy, and Ludkovski}{Binois
  et~al.}{2018}]{binois2018practical}
Binois, M., R.~B. Gramacy, and M.~Ludkovski. 2018.
\newblock ``Practical Heteroscedastic Gaussian Process Modeling for Large
  Simulation Experiments''.
\newblock {\em Journal of Computational and Graphical
  Statistics\/}~27(4):808--821.


\bibitem[\protect\citeauthoryear{Daulton, Balandat, and Bakshy}{Daulton
  et~al.}{2020}]{daulton2020ehvi}
Daulton, S., M.~Balandat, and E.~Bakshy. 2020.
\newblock ``Differentiable Expected Hypervolume Improvement for Parallel
  Multi-Objective {B}ayesian Optimization''.
\newblock In {\em Advances in Neural Information Processing Systems}, edited
  by\ H.~Larochelle, M.~Ranzato, R.~Hadsell, M.~Balcan, and H.~Lin.
\newblock December 6\textsuperscript{th}-12\textsuperscript{th}, virtual,
  9851--9864.

\bibitem[\protect\citeauthoryear{Daulton, Eriksson, Balandat, and
  Bakshy}{Daulton et~al.}{2021}]{daulton2021multi}
Daulton, S., D.~Eriksson, M.~Balandat, and E.~Bakshy. 2021.
\newblock ``Multi-Objective Bayesian Optimization over High-Dimensional Search
  Spaces''.
\newblock {\em arXiv preprint arXiv:2109.10964\/}.
\newblock Accessed: Jan 30, 2022.


\bibitem[\protect\citeauthoryear{del Rosario, Rupp, Kim, Antono, and Ling}{del
  Rosario et~al.}{2020}]{delrosario2020frontier}
del Rosario, Z., M.~Rupp, Y.~Kim, E.~Antono, and J.~Ling. 2020.
\newblock ``Assessing the frontier: Active learning, model accuracy, and
  multi-objective candidate discovery and optimization''.
\newblock {\em The Journal of Chemical Physics\/}~153(2):024112.


\bibitem[\protect\citeauthoryear{{Emmerich}, {Deutz}, and
  {Klinkenberg}}{{Emmerich} et~al.}{2011}]{emmerich2011hvei}
{Emmerich}, M. T.~M., A.~H. {Deutz}, and J.~W. {Klinkenberg}. 2011.
\newblock ``Hypervolume-Based Expected Improvement: Monotonicity Properties and
  Exact Computation''.
\newblock In {\em 2011 IEEE Congress of Evolutionary Computation (CEC)}.
\newblock June 5\textsuperscript{th}-8\textsuperscript{th}, New Orleans, LA,
  CA, 2147-2154.

\bibitem[\protect\citeauthoryear{Fasshauer and McCourt}{Fasshauer and
  McCourt}{2015}]{FasshauerGE2015ws}
Fasshauer, G.~E., and M.~J. McCourt. 2015.
\newblock {\em Kernel-based Approximation Methods Using {MATLAB}}.
\newblock Singapore: World Scientific.


\bibitem[\protect\citeauthoryear{Forrester, Sobester, and Keane}{Forrester
  et~al.}{2008}]{forrester2008engineering}
Forrester, A., A.~Sobester, and A.~Keane. 2008.
\newblock {\em Engineering Design via Surrogate Modelling: A Practical Guide}.
\newblock New York: John Wiley \& Sons.


\bibitem[\protect\citeauthoryear{Gardner, Kusner, Xu, Weinberger, and
  Cunningham}{Gardner et~al.}{2014}]{gardner_boconstraints}
Gardner, J.~R., M.~J. Kusner, Z.~E. Xu, K.~Q. Weinberger, and J.~P. Cunningham.
  2014.
\newblock ``Bayesian Optimization with Inequality Constraints''.
\newblock In {\em Proceedings of the 31\textsuperscript{th} International
  Conference on Machine Learning ({ICML} 2014)}, Volume~32.
\newblock June 21\textsuperscript{st}-26\textsuperscript{th}, Beijing, China,
  937--945.

\bibitem[\protect\citeauthoryear{Gardner, Pleiss, Bindel, Weinberger, and
  Wilson}{Gardner et~al.}{2018}]{gardner2018gpytorch}
Gardner, J.~R., G.~Pleiss, D.~Bindel, K.~Q. Weinberger, and A.~G. Wilson. 2018.
\newblock ``GPyTorch: Blackbox Matrix-Matrix Gaussian Process Inference with
  GPU Acceleration''.
\newblock In {\em Advances in Neural Information Processing Systems}, edited
  by\ S.~Bengio, H.~Wallach, H.~Larochelle, K.~Grauman, N.~Cesa-Bianchi, and
  R.~Garnett.
\newblock Dec 3\textsuperscript{rd}–8\textsuperscript{th}, Montreal, Canada.

\bibitem[\protect\citeauthoryear{Garnett}{Garnett}{2022}]{garnett_bayesoptbook_2022}
Garnett, R. 2022.
\newblock {\em {Bayesian Optimization}}.
\newblock Cambridge, UK: Cambridge University Press.
\newblock in preparation.


\bibitem[\protect\citeauthoryear{Garnett, Krishnamurthy, Xiong, Schneider, and
  Mann}{Garnett et~al.}{2012}]{garnett2012active}
Garnett, R., Y.~Krishnamurthy, X.~Xiong, J.~Schneider, and R.~Mann. 2012.
\newblock ``Bayesian Optimal Active Search and Surveying''.
\newblock In {\em Proceedings of the 29\textsuperscript{th} International
  Conference on International Conference on Machine Learning ({ICML} 2012)}.
\newblock June 26\textsuperscript{th}–July 1\textsuperscript{st}, Edinburgh,
  Scotland, 843–850.

\bibitem[\protect\citeauthoryear{Giuliani, Wechsung, Cerfon, Stadler, and
  Landreman}{Giuliani et~al.}{2022}]{giuliani2020single}
Giuliani, A., F.~Wechsung, A.~Cerfon, G.~Stadler, and M.~Landreman. 2022.
\newblock ``Single-Stage Gradient-Based Stellarator Coil Design: Optimization
  for Near-Axis Quasi-Symmetry''.
\newblock {\em Journal of Computational Physics\/}~459:111147.


\bibitem[\protect\citeauthoryear{{GPy}}{{GPy}}{ 2012}]{gpy2014}
{GPy} since 2012.
\newblock ``{GPy}: A Gaussian process framework in python''.
\newblock \url{http://github.com/SheffieldML/GPy}.
\newblock accessed Jan 12\textsuperscript{th}, 2022.

\bibitem[\protect\citeauthoryear{Haghanifar, McCourt, Cheng, Wuenschell,
  Ohodnicki, and Leu}{Haghanifar et~al.}{2020}]{haghanifar2020}
Haghanifar, S., M.~McCourt, B.~Cheng, J.~Wuenschell, P.~Ohodnicki, and P.~W.
  Leu. 2020.
\newblock ``Discovering High-Performance Broadband and Broad Angle
  Antireflection Surfaces by Machine Learning''.
\newblock {\em Optica\/}~7(7):784--789.


\bibitem[\protect\citeauthoryear{Hutter, Hoos, and Leyton-Brown}{Hutter
  et~al.}{2011}]{hutter2011smbo}
Hutter, F., H.~H. Hoos, and K.~Leyton-Brown. 2011.
\newblock ``Sequential Model-Based Optimization for General Algorithm
  Configuration''.
\newblock In {\em International Conference on Learning and Intelligent
  Optimization}, edited by\ C.~A.~C. Coello.
\newblock Jan 17\textsuperscript{th}-21\textsuperscript{st}, Rome, Italy,
  507--523.

\bibitem[\protect\citeauthoryear{Jiang, Malkomes, Converse, Shofner, Moseley,
  and Garnett}{Jiang et~al.}{2017}]{jiang2017enas}
Jiang, S., G.~Malkomes, G.~Converse, A.~Shofner, B.~Moseley, and R.~Garnett.
  2017.
\newblock ``Efficient Nonmyopic Active Search''.
\newblock In {\em Proceedings of the 34\textsuperscript{th} International
  Conference on Machine Learning, ({ICML} 2017)}, edited by\ D.~Precup and
  Y.~W. Teh, Volume~70.
\newblock August 6\textsuperscript{th}-11\textsuperscript{th}, Sydney,
  Australia, 1714--1723.

\bibitem[\protect\citeauthoryear{Jones, Schonlau, and Welch}{Jones
  et~al.}{1998}]{jones1998ego}
Jones, D.~R., M.~Schonlau, and W.~J. Welch. 1998, Dec.
\newblock ``Efficient Global Optimization of Expensive Black-Box Functions''.
\newblock {\em Journal of Global Optimization\/}~13(4):455--492.


\bibitem[\protect\citeauthoryear{{Knowles}}{{Knowles}}{2006}]{knowles2006parego}
{Knowles}, J. 2006.
\newblock ``{ParEGO}: A Hybrid Algorithm with On-Line Landscape Approximation
  for Expensive Multiobjective Optimization Problems''.
\newblock {\em IEEE Transactions on Evolutionary Computation\/}~10(1):50--66.


\bibitem[\protect\citeauthoryear{Malkomes, Cheng, Lee, and Mc{C}ourt}{Malkomes
  et~al.}{2021}]{malkomes2021cas}
Malkomes, G., B.~Cheng, E.~H. Lee, and M.~Mc{C}ourt. 2021.
\newblock ``Beyond the Pareto Efficient Frontier: Constraint Active Search for
  Multiobjective Experimental Design''.
\newblock In {\em Proceedings of the 38\textsuperscript{th} International
  Conference on Machine Learning ({ICML} 2021)}, edited by\ M.~Meila and
  T.~Zhang, Volume 139.
\newblock July 18\textsuperscript{th}-24\textsuperscript{th}, virtual,
  7423--7434.

\bibitem[\protect\citeauthoryear{Mockus, Tiesis, and Zilinskas}{Mockus
  et~al.}{1978}]{Mockus1978ei}
Mockus, J., V.~Tiesis, and A.~Zilinskas. 1978.
\newblock ``The Application of {B}ayesian Methods for Seeking the Extremum''.
\newblock {\em Towards Global Optimization\/}~2(117-129):2.


\bibitem[\protect\citeauthoryear{Molesky, Lin, Piggott, Jin, Vuckovi{\'c}, and
  Rodriguez}{Molesky et~al.}{2018}]{molesky2018inverse}
Molesky, S., Z.~Lin, A.~Y. Piggott, W.~Jin, J.~Vuckovi{\'c}, and A.~W.
  Rodriguez. 2018.
\newblock ``Inverse Design in Nanophotonics''.
\newblock {\em Nature Photonics\/}~12(11):659--670.


\bibitem[\protect\citeauthoryear{Negoescu, Frazier, and Powell}{Negoescu
  et~al.}{2011}]{negoescu2011knowledge}
Negoescu, D.~M., P.~I. Frazier, and W.~B. Powell. 2011.
\newblock ``The Knowledge-Gradient Algorithm for Sequencing Experiments in Drug
  Discovery''.
\newblock {\em INFORMS Journal on Computing\/}~23(3):346--363.


\bibitem[\protect\citeauthoryear{Niederreiter}{Niederreiter}{1992}]{niederreiter1992random}
Niederreiter, H. 1992.
\newblock {\em Random Number Generation and Quasi-Monte Carlo Methods}.
\newblock Philadelphia, PA, USA: Society for Industrial and Applied
  Mathematics.


\bibitem[\protect\citeauthoryear{Rasmussen and Williams}{Rasmussen and
  Williams}{2005}]{Rasmussen_gpforml}
Rasmussen, C.~E., and C.~K.~I. Williams. 2005.
\newblock {\em Gaussian Processes for Machine Learning (Adaptive Computation
  and Machine Learning)}.
\newblock Cambridge, MA, USA: The MIT Press.


\bibitem[\protect\citeauthoryear{Say{\i}n}{Say{\i}n}{2000}]{sayin2000measuring}
Say{\i}n, S. 2000.
\newblock ``Measuring the Quality of Discrete Representations of Efficient Sets
  in Multiple Objective Mathematical Programming''.
\newblock {\em Mathematical Programming\/}~87(3):543--560.


\bibitem[\protect\citeauthoryear{Scott, Frazier, and Powell}{Scott
  et~al.}{2011}]{scott2011gpkg}
Scott, W., P.~Frazier, and W.~Powell. 2011.
\newblock ``The Correlated Knowledge Gradient for Simulation Optimization of
  Continuous Parameters using Gaussian Process Regression''.
\newblock {\em SIAM Journal on Optimization\/}~21(3):996--1026.


\bibitem[\protect\citeauthoryear{Spitzer~Jr}{Spitzer~Jr}{1958}]{spitzer1958stellarator}
Spitzer~Jr, L. 1958.
\newblock ``The Stellarator Concept''.
\newblock {\em The Physics of Fluids\/}~1(4):253--264.


\bibitem[\protect\citeauthoryear{Srinivas, Krause, Kakade, and Seeger}{Srinivas
  et~al.}{2010}]{srinivas2010gpucb}
Srinivas, N., A.~Krause, S.~Kakade, and M.~Seeger. 2010.
\newblock ``Gaussian Process Optimization in the Bandit Setting: No Regret and
  Experimental Design''.
\newblock In {\em Proceedings of the 27\textsuperscript{th} International
  Conference on International Conference on Machine Learning (ICML 2010)},
  edited by\ J.~F{\"u}rnkranz and T.~Joachims.
\newblock June 21\textsuperscript{st}-24\textsuperscript{th}, Haifa, Israel,
  1015--1022.

\bibitem[\protect\citeauthoryear{Szechtman and Y{\"u}cesan}{Szechtman and
  Y{\"u}cesan}{2008}]{szechtman2008new}
Szechtman, R., and E.~Y{\"u}cesan. 2008.
\newblock ``A New Perspective on Feasibility Determination''.
\newblock In {\em Proceedings of the 2008 Winter Simulation Conference}, edited
  by\ S.~Mason, R.~Hill, L.~M{\"o}nch, and O.~Rose,  273--280.
\newblock Piscataway, New Jersey: Institute of Electrical and Electronics
  Engineers, Inc.

\bibitem[\protect\citeauthoryear{Szechtman and Y{\"u}cesan}{Szechtman and
  Y{\"u}cesan}{2016}]{szechtman2016bayesian}
Szechtman, R., and E.~Y{\"u}cesan. 2016.
\newblock ``A Bayesian Approach to Feasibility Determination''.
\newblock In {\em Proceedings of the 2016 Winter Simulation Conference}, edited
  by\ T.~M. Roeder, P.~I. Frazier, R.~Szechtman, E.~Zhou, T.~Huschka, and S.~E.
  Chick,  782--790.
\newblock Piscataway, New Jersey: Institute of Electrical and Electronics
  Engineers, Inc.

\bibitem[\protect\citeauthoryear{Tanabe and Ishibuchi}{Tanabe and
  Ishibuchi}{2020}]{tanabe2020reproblems}
Tanabe, R., and H.~Ishibuchi. 2020.
\newblock ``An Easy-to-Use Real-World Multi-Objective Optimization Problem
  Suite''.
\newblock {\em Applied Soft Computing\/}~89:106078.


\bibitem[\protect\citeauthoryear{Villemonteix, Vazquez, and
  Walter}{Villemonteix et~al.}{2009}]{villemonteix2009information}
Villemonteix, J., E.~Vazquez, and E.~Walter. 2009, aug.
\newblock ``An Informational Approach to the Global Optimization of
  Expensive-to-Evaluate Functions''.
\newblock {\em Journal of Global Optimization\/}~44(4):509–534.


\end{thebibliography}

\section*{AUTHOR BIOGRAPHIES}

\noindent {\bf ERIC HANS LEE} is a research scientist at SigOpt. His interests largely lie in Gaussian process modeling and global optimization. His email address is \email{eric.lee@intel.com}.\\

\noindent {\bf BOLONG CHENG} is the technical lead of the research team at SigOpt. He works on productionizing Bayesian optimization, and more broadly, sequential decision making problems.  Currently, he is interested in applying sequential optimization techniques in scientific and engineering domains such as materials simulation and design. His email address is \email{harvey.cheng@intel.com}.\\

\noindent {\bf MICHAEL MCCOURT} is a senior principal engineer at Intel and the general manager of SigOpt.  His recent work has focused on sample-efficient methods for design, involving both optimization and search.  Additionally, he has worked on matrix computations and kernel methods in math and statistics.  His email address is \email{mccourt@sigopt.com}.\\

\end{document}